%% file: main.tex
\documentclass[10pt,twocolumn,letterpaper]{article}

\usepackage{cvpr}              


\definecolor{cvprblue}{rgb}{0.21,0.49,0.74}
\usepackage[pagebackref,breaklinks,colorlinks,allcolors=cvprblue]{hyperref}

\usepackage[linesnumbered,ruled,vlined]{algorithm2e}

\def\paperID{*****} 
\def\confName{CVPR}
\def\confYear{2025}

\title{ResidualDroppath: Enhancing Feature Reuse over Residual Connections}

\author{Sejik Park\\
Graduate School of AI, Korea Advanced Institute of Science and Technology (KAIST)\\
South Korea\\
{\tt\small sejik.park@kaist.ac.kr}
}

\begin{document}
\maketitle
\input{sec/0_abstract}

\input{sec/1_intro}
\input{sec/2_rw}

\input{sec/3_meth}
\input{sec/4_exp}
\input{sec/5_conc}
{
    \small
    \bibliographystyle{ieeenat_fullname}
    \bibliography{main}
}

\input{sec/X_suppl}

\end{document}

%% file: sec/0_abstract.tex
\begin{abstract}

Residual connections are one of the most important components in neural network architectures for mitigating the vanishing gradient problem and facilitating the training of much deeper networks. One possible explanation for how residual connections aid deeper network training is by promoting feature reuse. However, we identify and analyze the limitations of feature reuse with vanilla residual connections. To address these limitations, we propose modifications in training methods. Specifically, we provide an additional opportunity for the model to learn feature reuse with residual connections through two types of iterations during training. The first type of iteration involves using droppath, which enforces feature reuse by randomly dropping a subset of layers. The second type of iteration focuses on training the dropped parts of the model while freezing the undropped parts. As a result, the dropped parts learn in a way that encourages feature reuse, as the model relies on the undropped parts with feature reuse in mind. Overall, we demonstrated performance improvements in models with residual connections for image classification in certain cases.


\end{abstract}


%% file: sec/1_intro.tex
\section{Introduction}
\label{sec:intro}

\begin{figure*}[htbp]
    \centering
    \includegraphics[width=0.9\textwidth]{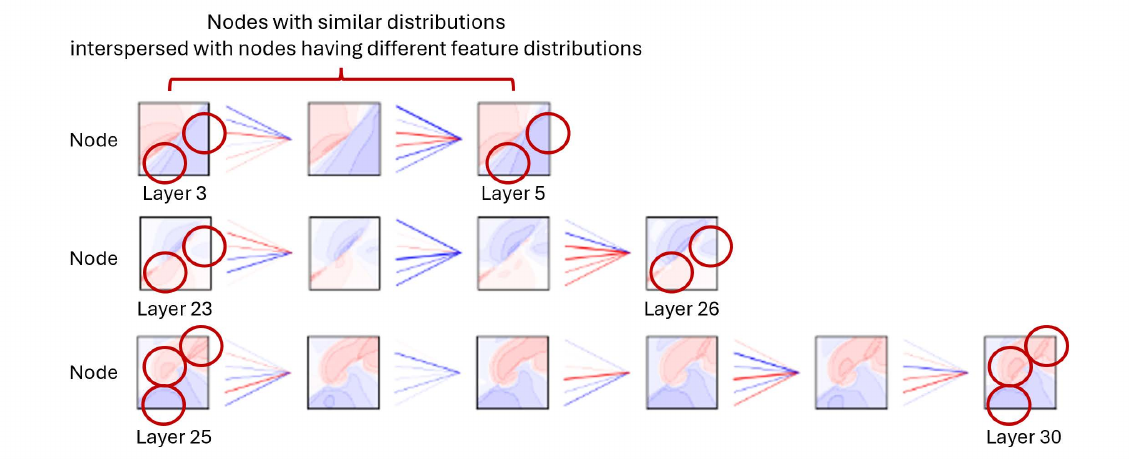}
    \caption{Feature Reuse Across Multiple Layers. It visualizes the feature distribution of a model trained with linear layers of depth 32 and a hidden dimension of 32, with residual connections added at each layer. Despite the presence of residual connections, it shows that the model produces similar feature distributions through multiple transformations of the previous layer's feature distribution. This could be disadvantageous from the perspective of information retention.}
    \label{fig:multiple}
\end{figure*}

Residual connection \cite{he2016deep} is such a general and popular technique that almost no model exists without it. For example, it is utilized in foundation models like GPT-4 \cite{openai2024gpt4technicalreport}, Llama 2 \cite{touvron2023llama}, and Stable Diffusion XL \cite{podell2023sdxlimprovinglatentdiffusion}. This is because residual connections allow gradients to flow more directly through layers, reducing the vanishing gradient problem. Additionally, the advantages of residual connections are sometimes interpreted as benefits of leveraging identity mapping across layers \cite{he2016identity}.

Our work shows that, despite the presence of residual connections, there are limitations in effectively utilizing identity mappings. To address this limitation, we propose an algorithm that provides additional opportunities for the model to learn identity mappings. This approach further enhances the model's performance.

The distinguishing feature of residual connections is their skip connection. This means that for a block within the model, the input \( x \) goes through a series of transformations \( F \) and then \( x \) is added back to produce the output. In other words, the operation \( F(x) + x \) is performed for the block. This can be seen in (a) of Figure~\ref{fig:algorithm}.

The residual connection utilizes the previous block's output directly with the skip connection, which can be interpreted as beneficial for feature reuse \cite{huang2017densely}. However, there is limited analysis within models on how features from previous blocks are reused. Considering the difficulty that deep learning models face when learning sparse matrices \cite{warmuth2020case}, it can be anticipated that reusing features across multiple blocks through residual connections may be challenging. Therefore, we analyzed the potential limitations of feature reuse in model training that includes residual connections.

To analyze feature reuse within the model, we visualized the node outputs directly. Specifically, we employed TFMeter~\cite{hoeiness2021tinkering}, an extension of TensorBoard Playground, which trains the model on a two-dimensional dataset and visualizes the output of nodes for grid inputs by generating contour plots. In this process, we scaled the analysis from the maximum of 6 layers supported by TFMeter to 32 layers. Through this, we demonstrated a deficiency in feature reuse across multiple layers, not only in small models but also in large models.

To address the deficiency, we propose an algorithm called ResidualDroppath. ResidualDroppath alternates between two types of iterations during training. In the first iteration, the model is trained with enforced feature reuse without any transformation over parts of the block by applying droppath \cite{huang2016deep}. In the second iteration, the model learns to consider whether to utilize these paths. This approach enables the model to reuse features in their identity form, facilitating feature reuse across multiple layers.

Then, we verified whether the proposed ResidualDroppath is beneficial for deep learning training in image classification tasks. To this end, we applied our algorithm to ResNet50 and ResNet50d models and tested it on the CIFAR10, MNIST, and ImageNet1k datasets. As a result, we observed a significant improvement in both Top-1 and Top-5 accuracy on the CIFAR10, and MNIST dataset. Furthermore, performance improvements were noted for the ResNet50d model on the ImageNet1K dataset.

Our contributions can be summarized as follows:
\begin{enumerate}
    \item We analyze the feature learning process and identify the challenges of reusing features across multiple layers.
    \item To address these limitations, we propose the ResidualDroppath algorithm, which enhances the model's ability to reuse features without transformation.
    \item We demonstrate that the proposed algorithm significantly improves the performance of ResNet50 and ResNet50d models on the CIFAR10 and MNIST image classification datasets. Additionally, on the ImageNet1K dataset, performance improvements were observed in ResNet50d.
\end{enumerate}


%% file: sec/2_rw.tex
\section{Related Work}
\label{sec:rw}

Our paper is broadly divided into an analysis of how the intermediate outputs of a model change depending on learning, and methods to enhance them. Therefore, we summarize related work on feature analysis in Section~\ref{sec:feature_analysis}. Additionally, we summarize the methods to enhance feature learning in Section~\ref{sec:enhance_feature}.

\subsection{Feature Analysis}
\label{sec:feature_analysis}
Recent works have conducted feature analysis by leveraging deep and high-dimensional neural networks to handle complex data. For instance, the feature output from intermediate layers is encoded using sparse autoencoders, and the correlation between these encoded values and the final output is analyzed \cite{templeton2024scaling, demircan2024sparse}. Another approach involves utilizing Singular Value Decomposition (SVD) to identify the presence of fine-grained concepts within the features \cite{graziani2023uncovering}. Alternatively, clustering features can create a dictionary of meaningful feature units \cite{dravid2023rosetta}. In contrast, we use a method where we train on a two-dimensional toy dataset and visualize the output on a grid input, plotting each dimension individually in two-dimensional space \cite{hoeiness2021tinkering}. In other words, we extend traditional visualization methods, originally intended for educational purposes, to more complex network architectures and learning algorithms. This approach provides an intuitive visualization of the features, which helps gain new insights into the critical aspects of the learning process. For additional feature analysis methods, please refer to Section~\ref{sec:suppl_related_analysis} of the Supplementary Materials.

\subsection{Enhancing Learning Feature}
\label{sec:enhance_feature}
The methods for influencing connections within the model to enhance performance based on the characteristics of learned features, included in PyTorch Image Models \cite{rw2019timm}, which is designed to reproduce a wide variety of state-of-the-art (SOTA) models, are as follows. First, there is a DenseNet method of connecting residual connections to all layers \cite{kim2024densenets, huang2017densely}. Next, there is the droppath method, where only the residual connection is retained while the computed values of the block are probabilistically masked \cite{oquab2023dinov2, huang2016deep}. Similarly, there are methods like dropout \cite{srivastava2014dropout} and dropblock \cite{ghiasi2018dropblock}, which apply masks to only a part of the block rather than the entire block. These methods can be regarded as approaches that allow the model to utilize short networks or thinned networks. On the other hand, our algorithm provides a different type of iteration in which one iteration references another to enhance the usage of residual connections. Other approaches that improve feature learning can be found in Supplementary Section~\ref{sec:suppl_related_enhance}.


%% file: sec/3_meth.tex
\section{Method}
\label{sec:meth}

In this paper, we propose an algorithm based on the analysis of features. Specifically, we analyze each dimension of the intermediate outputs of the model as features in Section~\ref{sec:characteristic}. Moreover, our feature analysis reveals that, even with residual connections, feature reuse across multiple layers is challenging. To address this limitation of feature reuse, we propose the ResidualDroppath algorithm in Section~\ref{sec:algorithm}.

\begin{figure}[t]
   \centering
   \includegraphics[width=0.4\linewidth]{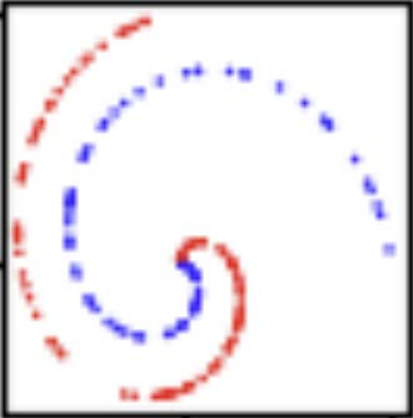}
   \caption{Toy Dataset. It visualizes 400 sampled points from the toy dataset based on the spiral function. Blue and red indicate the class of each point.}
   \label{fig:spiral}
\end{figure}

\subsection{Feature Reuse Across Multiple Layers}
\label{sec:characteristic} 
%
In this session, we demonstrate that naive residual connections may impede feature reuse across multiple layers. Specifically, when similar features appear in distant layers, we show that intermediate layers do not effectively leverage these features by transmitting them close to identity. This suggests that even when feature reuse is necessary, the model may tend to learn by applying additional transformations. Furthermore, we show that this learning approach may be disadvantageous in terms of information retention.

%
We utilized TFMeter~\cite{hoeiness2021tinkering}, an extension of the TensorBoard Playground, to analyze feature reuse. The primary distinction between the analysis conducted with TFMeter and our own lies in the network depth: while previous studies examined networks with a maximum of 6 layers, we extended this to a significantly deeper multi-layer perceptron (MLP) architecture, with up to 32 layers. For the training and test datasets, we adopted the following approach: a toy dataset containing 16,384 samples was generated for a binary classification task, with two classes based on a spiral function with 2D inputs. For visualization, the test data was designed with grid inputs, enabling us to observe not only the regions covered by the training data but also the potential range of values, including both in-distribution and out-of-distribution areas. The detailed overall process is as follows.

\begin{figure}[t]
   \centering
   \includegraphics[width=0.8\linewidth]{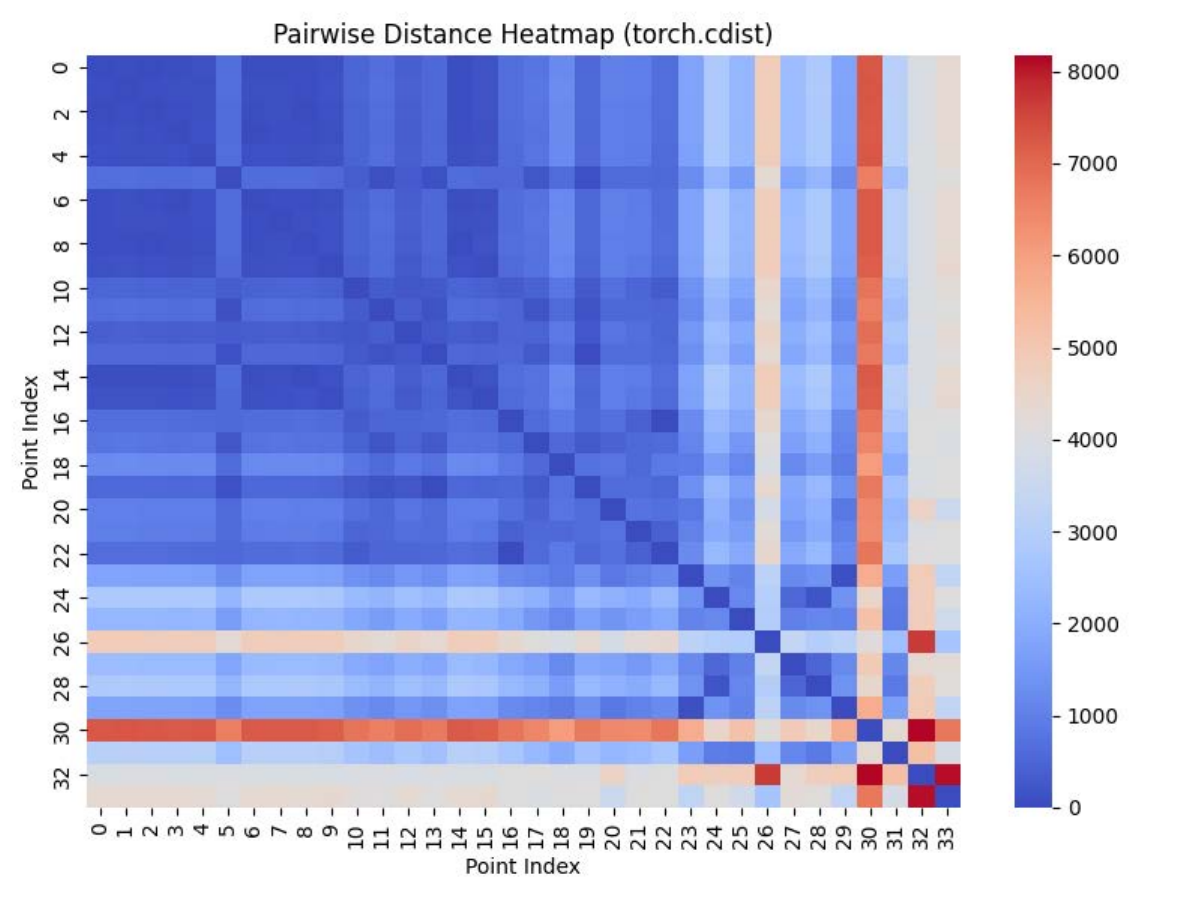}
   \caption{Feature Similarity over Layers with a Model of Depth 32 and Dimension 32. The similarity heatmap shows that similarity decreases and increases across multiple layers, which indirectly indicates that the model is performing multiple transformations of the previous layer’s feature distribution to obtain similar feature distributions.}
   \label{fig:duplicate_multi}
\end{figure}

%
The training data based on the spiral function can be visualized in Figure~\ref{fig:spiral}. Specifically, the data is generated in a form where the radius and angle are proportional to each other, and different labels are assigned by applying an angular shift. The detailed process is as follows. Angles $\alpha_1$ are sampled from a uniform distribution between $0$ and $2\pi$, which are used for label 1. The radius $rs$, proportional to these angles, is calculated using the formula $rs = \alpha / (2 * \pi)$, which is then used for both label 1 and label 2. The angles for label 2, denoted as shifted angles $\alpha_2$, are obtained using $\alpha_2 = \alpha_1 + \pi$. To summarize, the 2D data points and labels (x, y) are calculated as follows: for label 1, $(x, y) = (rs * \sin(\alpha_1), rs * \cos(\alpha_1)), 1)$, and for label 2, $(x, y) = (rs * \sin(\alpha_2), rs * \cos(\alpha_2)), 2)$.

%
To generate test data for visualization, grid samples are created with 50 equally spaced values along each axis within the range of [-1.0, 1.0]. These 2,500 samples encompass all coordinates covered by the training dataset. The node outputs corresponding to the grid samples are then visualized. Additionally, visualizations of previous weights are included to facilitate an understanding of transformations across layers.

The final visualization aims to illustrate \( N \) layers (which includes layer 1, layer 2, ..., up to layer \( N \)) with \( H \) hidden dimensions (which includes node 1, node 2, ..., up to node \( H \)), using \( H \times (2N+1) \) subplots. The first column visualizes the grid input corresponding to the input dimensions. In the even-numbered columns from 2 to \( 2N \), weights are represented as sequential lines, with color representing the sign and thickness representing the magnitude. In the odd-numbered columns from \( 2N+1 \), the output of the corresponding dimensions is visualized as a contour plot based on the grid input locations. For the final output, corresponding to \( 2N+1 \), the training data used is also displayed as a scatter plot. If the depth \( N \) and dimension \( H \) are large, a subset is sampled for visualization.

%
For visualization, the model and training hyperparameters were utilized as follows: We employed an n-layer MLP with GELU activation functions \cite{hendrycks2016gaussian} and residual connections in every layer. Two scales of models were used: a smaller model with a depth of 6 and hidden dimension of 6, and a larger model with a depth of 32 and hidden dimension of 32. The Adam base optimizer was employed for training \cite{kingma2014adam, loshchilov2017decoupled, dozat2016incorporating}, with the learning rate \(lr\) set to 0.1 and coefficients \(\beta\) set to (0.9, 0.999). Training was conducted over 1000 epochs with a batch size of 256, and each training run was carried out using an A6000 GPU.

\begin{figure}[t]
   \centering
   \includegraphics[width=0.8\linewidth]{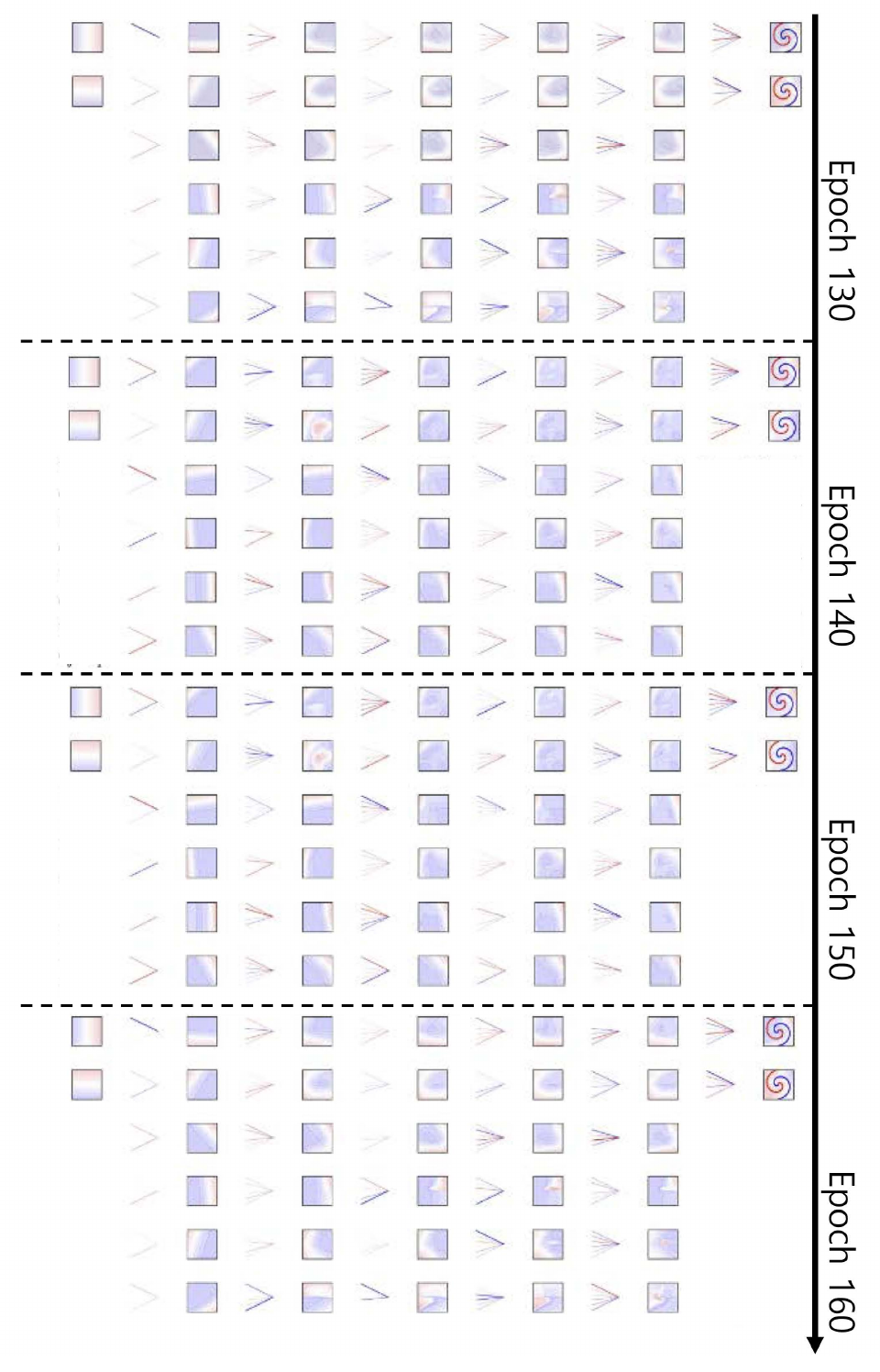}
   \caption{Feature Visualization during Training with a Model of Depth 6 and Dimension 6. It shows cases where some nodes achieve similar distributions across multiple layers.} 
   \label{fig:train}
\end{figure}

As a results, our visualization demonstrates that the model applies transformations across multiple layers to achieve similar feature distributions. For instance, as shown in Figure~\ref{fig:multiple}, certain nodes in a 32-layer model exhibit comparable distributions after passing through several layers, while intermediate layers display distinct distributions. Additionally, Figure~\ref{fig:duplicate_multi}, which illustrates feature similarity across each layer by calculating the Euclidean distance between the input and the features at all 32 layers using grid test data, supports this observation. Because the fluctuations in similarity observed across layers indirectly suggest that features with similar distributions emerge in non-consecutive layers. Moreover, the visualization of the 6-layer model across multiple epochs in Figure~\ref{fig:train} shows that this characteristic consistently appears during training.

%
And this phenomenon, in which each intermediate layer causes a distribution shift while multiple intermediate layers achieve a similar distribution in the first and last layers, can increase the potential for information loss. This is because the transformation induced by each layer may lead to a reduction in information \cite{wang2024yolov9}. Therefore, when feature reuse across different layers is necessary, it is essential to minimize distribution shifts across layers.

\begin{figure*}[htbp]
    \centering
    \includegraphics[width=0.9\textwidth]{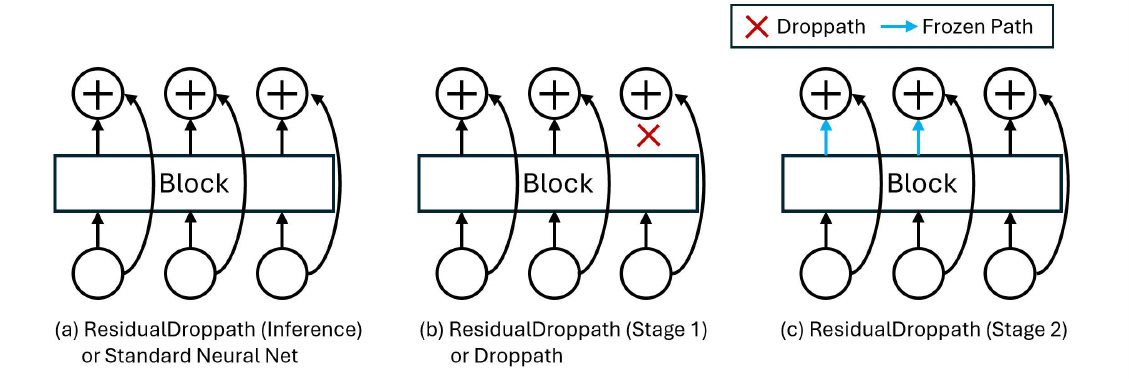}
    \caption{ResidualDroppath. It visualizes the operation of our proposed algorithm at the block level. There are two iteration stages. In the first stage, as shown in (b), Droppath is performed. In the second stage, paths that were not dropped in the first stage are frozen, while the remaining paths remain trainable. During inference, the process follows the conventional network flow, as illustrated in (a).}
    \label{fig:algorithm} 
\end{figure*}

\begin{algorithm}[t]
    \DontPrintSemicolon
	\caption{ResidualDroppath}
    \label{alg:residual}
    \textbf{Inputs:} Training dataset $D = \{(x_i, y_i)\}_{i=1}^{N}$ \\
    \textbf{Inputs:} Training Iteration $T$ \\
    \textbf{Inputs:} Pre-block $\theta_{pre}$, Block $\theta_1, ..., \theta_N$, Post-block $\theta_{post}$ \\
    \textbf{Inputs:} Batch size $B$, Loss function $L$ \\
    \textbf{Inputs:} Droppath stage $M$ \\
    $M = 0$ \\
    \For{$t=1,\dots,T$}{
        Sample $X, Y$ from $D$ \\
        $X \leftarrow \theta_{pre}(X)$ \\
        \For{$n=1,\dots,N$}{
            $X_{\text{step}} \leftarrow \theta_{n}(X)$ \\
            \If{\textcolor{red}{$M\%2 == 0$}}{
                $\text{mask}_d \leftarrow \text{sample\_mask}(X)$ \\ 
                $X \leftarrow X + X_{\text{step}} * \text{mask}_d$
            }
            \Else{
                \textcolor{red}{$X \leftarrow X + X_{\text{step}}\text{.detach()} * \text{mask}_d + X_{\text{step}} * (1-\text{mask}_d)$}
            }
        }
        $\hat{Y} \leftarrow \theta_{post}(X)$ \\
        Back-propagation with $L(Y, \hat{Y})$ \\
        $M \leftarrow M + 1$ \\
    }
\end{algorithm}

\subsection{Algorithm}
\label{sec:algorithm}
%
To reduce unnecessary transformations within the model that may cause a loss of information when learning to utilize features with similar distributions across different layers, we propose ResidualDroppath. This approach enhances feature reuse by iteratively alternating between two types of training iterations. The first type of iteration uses Droppath to enforce feature reuse by dropping certain residual parts, compelling the model to learn specifically in cases where feature reuse occurs. The second type of iteration references the first, centering the learning process on the dropped parts. This allows the model to learn whether to reuse features. The concept of this training algorithm is illustrated in Algorithm~\ref{alg:residual} or Figure~\ref{fig:algorithm}.

%
To summarize, as seen in Algorithm~\ref{alg:residual}, after an iteration using droppath (in black text), we focus on learning the parts dropped by the droppath (indicated in red text) in the subsequent step. In other words, building on the notion that Droppath \cite{huang2016deep} enforces the direct utilization of features, we introduce an additional iteration to learn from the enforced path, thereby enhancing the feature reuse capability of networks with residual connections. Unlike previous methods that rely on reversible functions or information-preserving losses \cite{wang2024yolov9}, our approach improves feature reuse with a two-stage iteration, guiding the model toward better information retention.


%% file: sec/4_exp.tex
\section{Experiments}
\label{sec:exp}


%
In this section, we evaluate the effectiveness of our ResidualDroppath on the image classification task. For the image classification task, we used the MNIST \cite{lecun1998gradient}, CIFAR10 \cite{krizhevsky2009learning}, and ImageNet1k \cite{imagenet15russakovsky} datasets. The model architectures used for applying the algorithm were ResNet50 \cite{he2016deep} and ResNet50d \cite{he2019bag}. Detailed explanations and results for each component are as follows.

\begin{table*}
  \centering
  \begin{tabular}{@{}ll|cc|cc|cc@{}}
    \toprule
    \multicolumn{2}{c|}{} & \multicolumn{2}{c|}{\textbf{MNIST}} & \multicolumn{2}{c|}{\textbf{CIFAR-10}} & \multicolumn{2}{c}{\textbf{Imagenet-1k}} \\
    \midrule
    Model & Algorithm & Top-1 & Top-5 & Top-1 & Top-5 & Top-1 & Top-5 \\
    \midrule
    ResNet50    & Standard  &  99.26 ± 0.05 &  \textbf{100.00 ± 0.00} &  88.07 ± 0.23 &  98.42 ± 0.09 &  75.52 & 91.89
 \\
                & Droppath  &  99.29 ± 0.03 &  99.99 ± 0.01          &  88.33 ± 0.12 &  98.61 ± 0.13 &  \textbf{76.11} & \textbf{92.49}
 \\
                & Ours      &  \textbf{99.32 ± 0.05} &  \textbf{100.00 ± 0.00} &  \textbf{89.22 ± 0.16} &  \textbf{98.87 ± 0.09} &  75.79 & 92.14 \\
    \midrule
    ResNet50d   & Standard  &  99.25 ± 0.07 &  \textbf{100.00 ± 0.00} &  89.55 ± 0.25 &  98.70 ± 0.09 &  76.16 & 92.40 \\
                & Droppath  &  99.27 ± 0.02 &  99.99 ± 0.00          &  89.84 ± 0.17 &  98.86 ± 0.08 &  76.49 & 92.77 \\
                & Ours      &  \textbf{99.32 ± 0.05} &  \textbf{100.00 ± 0.00} &  \textbf{90.84 ± 0.18} &  \textbf{98.94 ± 0.14} &  \textbf{76.57} & \textbf{92.78} \\
    \bottomrule
  \end{tabular}
  \caption{Accuracy for MNIST, CIFAR-10, and Imagenet-1k Datasets. Bold text indicates the highest mean accuracy. Each result was measured for test accuracy at the point when training was completed. The experiment for MNIST and CIFAR-10 was conducted over five different seeds. And the experiment for Imagenet-1k was conducted over one seed.}
  \label{tab:accuracy}
\end{table*}

\subsection{Settings}
Our experiment aimed to demonstrate that our algorithm performs well across various dataset scales by using three datasets with differing image sizes and quantities. The datasets used were as follows: First, we utilized MNIST, a grayscale dataset with an image size of (28, 28). This dataset consists of 70,000 images across 10 classes representing different digits, with 60,000 images used for training and 10,000 images for testing. Second, we used CIFAR-10, a color dataset with an image size of (32, 32). This dataset includes 60,000 images across 10 classes that distinguish objects and animals, with 50,000 images for training and 10,000 images for testing. Third, we employed ImageNet-1k, a color dataset with an image size of (224, 224), comprising approximately 1.3 million images across 1,000 classes of specific objects and animals. Of these, 1,281,167 images were used for training, and 50,000 images were allocated as a validation dataset, which we also used as the test dataset.

To ensure training within a reasonable time frame, given the differing scales of the datasets, we adjusted the batch size accordingly. We used a batch size of 128 to train MNIST and CIFAR-10 on a single A6000 GPU. For ImageNet-1k, we used a batch size of 1024 and trained on four A100 GPUs.

Then, we aimed to demonstrate that our algorithm provides significant performance improvements when applied to model architectures that leverage residual connections. For this purpose, we used ResNet50 and its enhanced version, ResNet50d, with hyperparameters set to the default values provided in PyTorch Image Models \cite{rw2019timm}.

The baseline comparison was conducted using a model trained with a standard training procedure (without any additional procedure like Droppath or ResidaulDroppath) and a model trained with the Droppath training procedure \cite{huang2016deep}. For Droppath, all hyperparameters were set identically to the default settings in PyTorch Image Models \cite{rw2019timm}, except for one. Specifically, the drop rate was set to 0.1. The one difference in settings relates to maintaining the distribution scale of the residual output. In Droppath, a scaling factor is applied based on the drop rate to preserve the distribution of the residuals as they would be without Droppath. However, in our ResidualDroppath, the dropped part of the residual distribution should be treated as a zero matrix, so scaling was not applied.

\subsection{Results}
As shown in Table~\ref{tab:accuracy}, our method demonstrates superior top-1 and top-5 accuracy compared to other methods when information is richer for each class. This indicates that our algorithm effectively helps retain and leverage the increased information for learning. In other words, this highlights the effectiveness of our Droppath algorithm in enhancing the performance of models with residual connections by facilitating feature reuse. A comprehensive analysis is provided below.

Our method provides a slight performance edge on the MNIST dataset. Specifically, our algorithm achieves a top-1 accuracy of 99.32\% with the ResNet50 model, showing an improvement of 0.06\% and 0.03\% over the Standard and Droppath methods, respectively. Similarly, with the ResNet50d model, our approach demonstrates a performance improvement of 0.07\% and 0.05\% compared to the Standard and Droppath baselines.

Our algorithm shows a greater performance difference compared to other methods on the CIFAR-10 dataset than on the MNIST dataset. This may be due to its effective utilization of newly added color information, which is not present in MNIST. Detailed results are as follows: our algorithm achieves a Top-1 accuracy of 90.84\% on the ResNet50d model. This demonstrates a performance improvement of over 1\% in Top-1 accuracy compared to other methods, such as Standard and Droppath. Similarly, enhanced results are observed on the ResNet50 model in both Top-1 and Top-5 accuracies.

Our algorithm demonstrated the best performance on the ImageNet-1K dataset with ResNet50d, although it achieved slightly lower performance than Droppath when applied to ResNet50. This may be attributed to a reduction in information per class as the number of images per class decreases. The detailed results are as follows: For ResNet50d, our algorithm achieved a Top-1 accuracy of 76.57\%, marking the highest performance. In terms of Top-5 accuracy, our model showed a gain of 0.38\% over the standard model and 0.01\% over Droppath. However, for ResNet50, we observed a decrease in Top-1 accuracy by 0.32\%.


%% file: sec/5_conc.tex
\section{Conclusion}
\label{sec:conc}


%
With similarity analysis of layer-wise features on a toy dataset, we demonstrate that residual connections in existing methods are inefficient in preserving and utilizing information across multiple layers. To address this, we alternate between two-stage iterations: enforcing full information preservation through a masked residual connection and learning the enforced residual components. As a result, our proposed algorithm improves performance on various image classification datasets using ResNet50 and ResNet50d.

%
Our limitation is that ResNet-50 with our algorithm does not achieve the best performance on ImageNet-1K. We interpreted this as potentially due to differences in the amount of information across classes, suggesting the need for a more detailed analysis of this factor. Furthermore, our method does not account for the possibility that, as illustrated in Feature~\ref{fig:train}, multiple nodes within the same layer may exhibit similar feature distributions. Although our approach assumes a distribution shift across intermediate layers when using the same features across multiple layers, the actual model training process may involve complex inter-node relationships that allow for more efficient feature reuse.

%
For future work, we plan to investigate how the model can retain features within each layer to enable effective feature reuse. As shown in Figure~\ref{fig:multiple} or Figure~\ref{fig:train}, transformations occur between features of different layers through weights. This implies that, to achieve feature reuse, certain nodes may need opposing signs to allow for cancellation, or conditions where all weights are effectively zero. In summary, exploring the characteristics of features that are useful for the model could be a valuable research direction. Additionally, for models with residual connections, architectures beyond ResNet, such as transformers, are worth considering.


%% file: sec/X_suppl.tex
\clearpage
\setcounter{page}{1}
\maketitlesupplementary

\section{Supplementary}
\label{sec:suppl}

\subsection{Acknowledgments}
We appreciate the high-performance GPU computing support of HPC-AI Open Infrastructure via GIST SCENT.

\subsection{Related Work: Feature Analysis}
\label{sec:suppl_related_analysis}

There are several methods to analyze features: directly analyzing features, analyzing the relationships between features, or analyzing the differences that arise from modifying features. Direct methods involve analyzing features, including the output, which has been addressed in various studies \cite{staats2024locating, templeton2024scaling, graziani2023uncovering, dravid2023rosetta, oh2024house, huh2024platonic, song2024out, kim2024mechanistic, lv2024language, vesdapunt2024hvclip, attiasinformation, uchiyama2024programminglanguagefeaturespretraining, liang2022mind, zhang2024connect, he2024debiasing}. When considering relationships, methods can involve examining the relationship between input and features \cite{bau2017network, bau2018gan}, between output and features \cite{song2024out, huh2024platonic, bandarkar2024layer, zhang2024co, yang2023dynamic}, or the relationship between input and output \cite{attiasinformation}. Methods that analyze differences resulting from changes include modifying features and observing the change in output \cite{jiang2024comparing, ghiasi2022vision, liang2024parameterefficientfinetuningspectraldomain, park2024diverse, park2024learning, wang2023efficienttrain, kaddour2024no, li2022automated, burda2018exploration, zoph2020learning, cubuk2020randaugment, hendrycks2019augmix, touvron2022deit, zhang2017mixup, yun2019cutmix, dravid2023rosetta, zeiler2014visualizing, olah2020zoom}, using only a subset of features to observe the differences \cite{bhaskar2024heuristic}, or modifying the input to see the changes in output \cite{jiang2024comparing}. Feature analysis can be performed within a single model \cite{lv2024language, xu2024collapsedlanguagemodelspromote, donahue2014decaf}, or across different models \cite{liang2022mind, zhang2024connect, ghiasi2022vision, anonymous2024layer}. The visualization-based analysis we conducted falls under the category of direct feature analysis within a single model.

\subsection{Related Work: Enhancing Learning Feature}
\label{sec:suppl_related_enhance}
To improve feature learning, appropriate feature exploration should be conducted to allow for the learning of compositional features \cite{du2024compositional} \cite{park2024diverse, park2024learning}. In other words, learning should proceed with exploration that allows for exploitation in meaningful regions. The following approaches can be taken to achieve this goal: either directly increasing the exploration space by providing diverse features to the model through augmentation and other techniques \cite{sakurai2024lare, liu2023soft, wang2023image, park2024diverse, park2024learning, wang2023efficienttrain, kaddour2024no, li2022automated, burda2018exploration, zoph2020learning, cubuk2020randaugment, hendrycks2019augmix, touvron2022deit, zhang2017mixup, yun2019cutmix, vesdapunt2024hvclip, brief2024mixing, li2024formalityfavoredunravelinglearning, grangier2024taskadaptivepretrainedlanguagemodels, liang2024parameterefficientfinetuningspectraldomain}, or indirectly designing a model that can learn the characteristics of the data \cite{xia2403vit, guo2024cpp, zhai2023feature, qiu2023cafeboost, attaiki2023understanding, potje2023enhancing, yu2023adaptive, khan2023localized, edstedt2024roma, hong2024concept, duan2024content, springenberg2014striving, xiao2024improving, xie2024making, darcet2023vision, han2024slimletllmlearn, he2024upcyclinglargelanguagemodels, luo2024monointernvlpushingboundariesmonolithic, qiu2024unlocking, cai2024adashift, tang2024smile, sun2024svfit, gao2024enhancing, simon2020adaptive, lizzo2024unlearn, son2024sg, gaya2022building, liu2024pat, zhao2024enhancing, li2024robustness, samragh2024scaling, lin2023class, serra2018overcoming, kim2024mechanistic, bandarkar2024layer, kim2023solar}. Alternatively, the loss function can be adjusted to better align with the learning objectives \cite{kim2023feature, lee2023fix, ding2024finetuningstrugglesforgettingmachine, ma2023curvature, cao2024sorsa, ni2024pace, baidoes, barcelo2024avoiding, cho2024dsg, li2020adversarial}. Another approach involves reducing the feature space that can be learned in a controlled manner using regularization \cite{wang2023learning, tran2024preserving, oh2024house, gu2023preserving, foret2021sharpnessaware, wang2024yolov9, tseng2020regularizing, zhang2020deepemd, kobayashi2024mean, wu2024panadapter, zheng2024learning, yang2024graphlora, xu2024collapsedlanguagemodelspromote, donahue2014decaf, feng2024tasl, mentzer2023finite, esser2024scaling, liu2024zero, yu2023block}. Our approach adds a regularization that utilizes residual connections to help feature exploration during learning.


%% file: main.bbl
\begin{thebibliography}{131}
\providecommand{\natexlab}[1]{#1}
\providecommand{\url}[1]{\texttt{#1}}
\expandafter\ifx\csname urlstyle\endcsname\relax
  \providecommand{\doi}[1]{doi: #1}\else
  \providecommand{\doi}{doi: \begingroup \urlstyle{rm}\Url}\fi

\bibitem[Anonymous(2024)]{anonymous2024layer}
Anonymous.
\newblock Layer by layer: Uncovering where multi-task learning happens in instruction-tuned large language models.
\newblock In \emph{Submitted to ACL Rolling Review - June 2024}, 2024.
\newblock under review.

\bibitem[Attaiki and Ovsjanikov(2023)]{attaiki2023understanding}
Souhaib Attaiki and Maks Ovsjanikov.
\newblock Understanding and improving features learned in deep functional maps.
\newblock In \emph{Proceedings of the IEEE/CVF Conference on Computer Vision and Pattern Recognition}, pages 1316--1326, 2023.

\bibitem[Attias et~al.()Attias, Dziugaite, Haghifam, Livni, and Roy]{attiasinformation}
Idan Attias, Gintare~Karolina Dziugaite, Mahdi Haghifam, Roi Livni, and Daniel~M Roy.
\newblock Information complexity of stochastic convex optimization: Applications to generalization, memorization, and tracing.
\newblock In \emph{Forty-first International Conference on Machine Learning}.

\bibitem[Bai et~al.()Bai, Sun, and Balasubramanian]{baidoes}
Xueying Bai, Yifan Sun, and Niranjan Balasubramanian.
\newblock Does roberta perform better than bert in continual learning: An attention sink perspective.
\newblock In \emph{First Conference on Language Modeling}.

\bibitem[Bandarkar et~al.(2024)Bandarkar, Muller, Yuvraj, Hou, Singhal, Lv, and Liu]{bandarkar2024layer}
Lucas Bandarkar, Benjamin Muller, Pritish Yuvraj, Rui Hou, Nayan Singhal, Hongjiang Lv, and Bing Liu.
\newblock Layer swapping for zero-shot cross-lingual transfer in large language models.
\newblock \emph{arXiv preprint arXiv:2410.01335}, 2024.

\bibitem[Barcel{\'o} et~al.(2024)Barcel{\'o}, Alc{\'a}zar, and Tobar]{barcelo2024avoiding}
Roberto Barcel{\'o}, Crist{\'o}bal Alc{\'a}zar, and Felipe Tobar.
\newblock Avoiding mode collapse in diffusion models fine-tuned with reinforcement learning.
\newblock \emph{arXiv preprint arXiv:2410.08315}, 2024.

\bibitem[Bau et~al.(2017)Bau, Zhou, Khosla, Oliva, and Torralba]{bau2017network}
David Bau, Bolei Zhou, Aditya Khosla, Aude Oliva, and Antonio Torralba.
\newblock Network dissection: Quantifying interpretability of deep visual representations.
\newblock In \emph{Proceedings of the IEEE conference on computer vision and pattern recognition}, pages 6541--6549, 2017.

\bibitem[Bau et~al.(2018)Bau, Zhu, Strobelt, Zhou, Tenenbaum, Freeman, and Torralba]{bau2018gan}
David Bau, Jun-Yan Zhu, Hendrik Strobelt, Bolei Zhou, Joshua~B Tenenbaum, William~T Freeman, and Antonio Torralba.
\newblock Gan dissection: Visualizing and understanding generative adversarial networks.
\newblock \emph{arXiv preprint arXiv:1811.10597}, 2018.

\bibitem[Bhaskar et~al.(2024)Bhaskar, Friedman, and Chen]{bhaskar2024heuristic}
Adithya Bhaskar, Dan Friedman, and Danqi Chen.
\newblock The heuristic core: Understanding subnetwork generalization in pretrained language models.
\newblock \emph{arXiv preprint arXiv:2403.03942}, 2024.

\bibitem[Brief et~al.(2024)Brief, Ovadia, Shenderovitz, Yoash, Lemberg, and Sheetrit]{brief2024mixing}
Meni Brief, Oded Ovadia, Gil Shenderovitz, Noga~Ben Yoash, Rachel Lemberg, and Eitam Sheetrit.
\newblock Mixing it up: The cocktail effect of multi-task fine-tuning on llm performance--a case study in finance.
\newblock \emph{arXiv preprint arXiv:2410.01109}, 2024.

\bibitem[Burda et~al.(2018)Burda, Edwards, Storkey, and Klimov]{burda2018exploration}
Yuri Burda, Harrison Edwards, Amos Storkey, and Oleg Klimov.
\newblock Exploration by random network distillation.
\newblock \emph{arXiv preprint arXiv:1810.12894}, 2018.

\bibitem[Cai(2024)]{cai2024adashift}
Sudong Cai.
\newblock Adashift: Learning discriminative self-gated neural feature activation with an adaptive shift factor.
\newblock In \emph{Proceedings of the IEEE/CVF Conference on Computer Vision and Pattern Recognition}, pages 5947--5956, 2024.

\bibitem[Cao(2024)]{cao2024sorsa}
Yang Cao.
\newblock Sorsa: Singular values and orthonormal regularized singular vectors adaptation of large language models.
\newblock \emph{arXiv preprint arXiv:2409.00055}, 2024.

\bibitem[Cho et~al.(2024)Cho, Jeon, Lee, Lee, and Kim]{cho2024dsg}
Sangyeon Cho, Jangyeong Jeon, Dongjoon Lee, Changhee Lee, and Junyeong Kim.
\newblock Dsg-kd: Knowledge distillation from domain-specific to general language models.
\newblock \emph{IEEE Access}, 2024.

\bibitem[Cubuk et~al.(2020)Cubuk, Zoph, Shlens, and Le]{cubuk2020randaugment}
Ekin~D Cubuk, Barret Zoph, Jonathon Shlens, and Quoc~V Le.
\newblock Randaugment: Practical automated data augmentation with a reduced search space.
\newblock In \emph{Proceedings of the IEEE/CVF conference on computer vision and pattern recognition workshops}, pages 702--703, 2020.

\bibitem[Darcet et~al.(2023)Darcet, Oquab, Mairal, and Bojanowski]{darcet2023vision}
Timoth{\'e}e Darcet, Maxime Oquab, Julien Mairal, and Piotr Bojanowski.
\newblock Vision transformers need registers.
\newblock \emph{arXiv preprint arXiv:2309.16588}, 2023.

\bibitem[Demircan et~al.(2024)Demircan, Saanum, Jagadish, Binz, and Schulz]{demircan2024sparse}
Can Demircan, Tankred Saanum, Akshay~K Jagadish, Marcel Binz, and Eric Schulz.
\newblock Sparse autoencoders reveal temporal difference learning in large language models.
\newblock \emph{arXiv preprint arXiv:2410.01280}, 2024.

\bibitem[Ding et~al.(2024)Ding, Xu, and Ji]{ding2024finetuningstrugglesforgettingmachine}
Meng Ding, Jinhui Xu, and Kaiyi Ji.
\newblock Why fine-tuning struggles with forgetting in machine unlearning? theoretical insights and a remedial approach, 2024.

\bibitem[Donahue et~al.(2014)Donahue, Jia, Vinyals, Hoffman, Zhang, Tzeng, and Darrell]{donahue2014decaf}
Jeff Donahue, Yangqing Jia, Oriol Vinyals, Judy Hoffman, Ning Zhang, Eric Tzeng, and Trevor Darrell.
\newblock Decaf: A deep convolutional activation feature for generic visual recognition.
\newblock In \emph{International conference on machine learning}, pages 647--655. PMLR, 2014.

\bibitem[Dozat(2016)]{dozat2016incorporating}
Timothy Dozat.
\newblock Incorporating nesterov momentum into adam.
\newblock 2016.

\bibitem[Dravid et~al.(2023)Dravid, Gandelsman, Efros, and Shocher]{dravid2023rosetta}
Amil Dravid, Yossi Gandelsman, Alexei~A Efros, and Assaf Shocher.
\newblock Rosetta neurons: Mining the common units in a model zoo.
\newblock In \emph{Proceedings of the IEEE/CVF International Conference on Computer Vision}, pages 1934--1943, 2023.

\bibitem[Du and Kaelbling(2024)]{du2024compositional}
Yilun Du and Leslie Kaelbling.
\newblock Compositional generative modeling: A single model is not all you need.
\newblock \emph{arXiv preprint arXiv:2402.01103}, 2024.

\bibitem[Duan et~al.(2024)Duan, Wu, Deng, and Deng]{duan2024content}
Yule Duan, Xiao Wu, Haoyu Deng, and Liang-Jian Deng.
\newblock Content-adaptive non-local convolution for remote sensing pansharpening.
\newblock In \emph{Proceedings of the IEEE/CVF Conference on Computer Vision and Pattern Recognition}, pages 27738--27747, 2024.

\bibitem[Edstedt et~al.(2024)Edstedt, Sun, B{\"o}kman, Wadenb{\"a}ck, and Felsberg]{edstedt2024roma}
Johan Edstedt, Qiyu Sun, Georg B{\"o}kman, M{\aa}rten Wadenb{\"a}ck, and Michael Felsberg.
\newblock Roma: Robust dense feature matching.
\newblock In \emph{Proceedings of the IEEE/CVF Conference on Computer Vision and Pattern Recognition}, pages 19790--19800, 2024.

\bibitem[Esser et~al.(2024)Esser, Kulal, Blattmann, Entezari, M{\"u}ller, Saini, Levi, Lorenz, Sauer, Boesel, et~al.]{esser2024scaling}
Patrick Esser, Sumith Kulal, Andreas Blattmann, Rahim Entezari, Jonas M{\"u}ller, Harry Saini, Yam Levi, Dominik Lorenz, Axel Sauer, Frederic Boesel, et~al.
\newblock Scaling rectified flow transformers for high-resolution image synthesis.
\newblock In \emph{Forty-first International Conference on Machine Learning}, 2024.

\bibitem[Feng et~al.(2024)Feng, Chu, Xu, Lu, Liu, Yu, and Wu]{feng2024tasl}
Yujie Feng, Xu Chu, Yongxin Xu, Zexin Lu, Bo Liu, Philip~S Yu, and Xiao-Ming Wu.
\newblock Tasl: Task skill localization and consolidation for language model continual learning.
\newblock \emph{arXiv preprint arXiv:2408.05200}, 2024.

\bibitem[Foret et~al.(2021)Foret, Kleiner, Mobahi, and Neyshabur]{foret2021sharpnessaware}
Pierre Foret, Ariel Kleiner, Hossein Mobahi, and Behnam Neyshabur.
\newblock Sharpness-aware minimization for efficiently improving generalization.
\newblock In \emph{International Conference on Learning Representations}, 2021.

\bibitem[Gao et~al.(2024)Gao, Zhang, Wu, and Li]{gao2024enhancing}
Jian Gao, Xiao Zhang, Ji Wu, and Miao Li.
\newblock Enhancing elusive clues in knowledge learning by contrasting attention of language models.
\newblock \emph{arXiv preprint arXiv:2409.17954}, 2024.

\bibitem[Gaya et~al.(2022)Gaya, Doan, Caccia, Soulier, Denoyer, and Raileanu]{gaya2022building}
Jean-Baptiste Gaya, Thang Doan, Lucas Caccia, Laure Soulier, Ludovic Denoyer, and Roberta Raileanu.
\newblock Building a subspace of policies for scalable continual learning.
\newblock \emph{arXiv preprint arXiv:2211.10445}, 2022.

\bibitem[Ghiasi et~al.(2022)Ghiasi, Kazemi, Borgnia, Reich, Shu, Goldblum, Wilson, and Goldstein]{ghiasi2022vision}
Amin Ghiasi, Hamid Kazemi, Eitan Borgnia, Steven Reich, Manli Shu, Micah Goldblum, Andrew~Gordon Wilson, and Tom Goldstein.
\newblock What do vision transformers learn? a visual exploration.
\newblock \emph{arXiv preprint arXiv:2212.06727}, 2022.

\bibitem[Ghiasi et~al.(2018)Ghiasi, Lin, and Le]{ghiasi2018dropblock}
Golnaz Ghiasi, Tsung-Yi Lin, and Quoc~V Le.
\newblock Dropblock: A regularization method for convolutional networks.
\newblock \emph{Advances in neural information processing systems}, 31, 2018.

\bibitem[Grangier et~al.(2024)Grangier, Fan, Seto, and Ablin]{grangier2024taskadaptivepretrainedlanguagemodels}
David Grangier, Simin Fan, Skyler Seto, and Pierre Ablin.
\newblock Task-adaptive pretrained language models via clustered-importance sampling, 2024.

\bibitem[Graziani et~al.(2023)Graziani, Mahony, Nguyen, M{\"u}ller, and Andrearczyk]{graziani2023uncovering}
Mara Graziani, Laura~O' Mahony, An-Phi Nguyen, Henning M{\"u}ller, and Vincent Andrearczyk.
\newblock Uncovering unique concept vectors through latent space decomposition.
\newblock \emph{arXiv preprint arXiv:2307.06913}, 2023.

\bibitem[Gu et~al.(2023)Gu, Shim, and Shkurti]{gu2023preserving}
Qiao Gu, Dongsub Shim, and Florian Shkurti.
\newblock Preserving linear separability in continual learning by backward feature projection.
\newblock In \emph{Proceedings of the IEEE/CVF Conference on Computer Vision and Pattern Recognition}, pages 24286--24295, 2023.

\bibitem[Guo and Gan(2024)]{guo2024cpp}
Zhen Guo and Hongping Gan.
\newblock Cpp-net: Embracing multi-scale feature fusion into deep unfolding cp-ppa network for compressive sensing.
\newblock In \emph{Proceedings of the IEEE/CVF Conference on Computer Vision and Pattern Recognition}, pages 25086--25095, 2024.

\bibitem[Han et~al.(2024)Han, Du, Du, Zhou, Wu, Zheng, and Han]{han2024slimletllmlearn}
Jiayi Han, Liang Du, Hongwei Du, Xiangguo Zhou, Yiwen Wu, Weibo Zheng, and Donghong Han.
\newblock Slim: Let llm learn more and forget less with soft lora and identity mixture, 2024.

\bibitem[He et~al.(2024{\natexlab{a}})He, Khattar, Prenger, Korthikanti, Yan, Liu, Fan, Aithal, Shoeybi, and Catanzaro]{he2024upcyclinglargelanguagemodels}
Ethan He, Abhinav Khattar, Ryan Prenger, Vijay Korthikanti, Zijie Yan, Tong Liu, Shiqing Fan, Ashwath Aithal, Mohammad Shoeybi, and Bryan Catanzaro.
\newblock Upcycling large language models into mixture of experts, 2024{\natexlab{a}}.

\bibitem[He et~al.(2016{\natexlab{a}})He, Zhang, Ren, and Sun]{he2016deep}
Kaiming He, Xiangyu Zhang, Shaoqing Ren, and Jian Sun.
\newblock Deep residual learning for image recognition.
\newblock In \emph{Proceedings of the IEEE conference on computer vision and pattern recognition}, pages 770--778, 2016{\natexlab{a}}.

\bibitem[He et~al.(2016{\natexlab{b}})He, Zhang, Ren, and Sun]{he2016identity}
Kaiming He, Xiangyu Zhang, Shaoqing Ren, and Jian Sun.
\newblock Identity mappings in deep residual networks.
\newblock In \emph{Computer Vision--ECCV 2016: 14th European Conference, Amsterdam, The Netherlands, October 11--14, 2016, Proceedings, Part IV 14}, pages 630--645. Springer, 2016{\natexlab{b}}.

\bibitem[He et~al.(2024{\natexlab{b}})He, Xue, Tan, Zhang, Yu, Bai, and Qi]{he2024debiasing}
Ruifei He, Chuhui Xue, Haoru Tan, Wenqing Zhang, Yingchen Yu, Song Bai, and Xiaojuan Qi.
\newblock Debiasing text-to-image diffusion models.
\newblock \emph{arXiv preprint arXiv:2402.14577}, 2024{\natexlab{b}}.

\bibitem[He et~al.(2019)He, Zhang, Zhang, Zhang, Xie, and Li]{he2019bag}
Tong He, Zhi Zhang, Hang Zhang, Zhongyue Zhang, Junyuan Xie, and Mu Li.
\newblock Bag of tricks for image classification with convolutional neural networks.
\newblock In \emph{Proceedings of the IEEE/CVF conference on computer vision and pattern recognition}, pages 558--567, 2019.

\bibitem[Hendrycks and Gimpel(2016)]{hendrycks2016gaussian}
Dan Hendrycks and Kevin Gimpel.
\newblock Gaussian error linear units (gelus).
\newblock \emph{arXiv preprint arXiv:1606.08415}, 2016.

\bibitem[Hendrycks et~al.(2019)Hendrycks, Mu, Cubuk, Zoph, Gilmer, and Lakshminarayanan]{hendrycks2019augmix}
Dan Hendrycks, Norman Mu, Ekin~D Cubuk, Barret Zoph, Justin Gilmer, and Balaji Lakshminarayanan.
\newblock Augmix: A simple data processing method to improve robustness and uncertainty.
\newblock \emph{arXiv preprint arXiv:1912.02781}, 2019.

\bibitem[Hoeiness et~al.(2021)Hoeiness, Harstad, and Friedland]{hoeiness2021tinkering}
Henrik Hoeiness, Axel Harstad, and Gerald Friedland.
\newblock From tinkering to engineering: Measurements in tensorflow playground.
\newblock \emph{arXiv preprint arXiv:2101.04141}, 2021.

\bibitem[Hong et~al.(2024)Hong, Park, and Pavlic]{hong2024concept}
Jinyung Hong, Keun~Hee Park, and Theodore~P Pavlic.
\newblock Concept-centric transformers: Enhancing model interpretability through object-centric concept learning within a shared global workspace.
\newblock In \emph{Proceedings of the IEEE/CVF Winter Conference on Applications of Computer Vision}, pages 4880--4891, 2024.

\bibitem[Huang et~al.(2016)Huang, Sun, Liu, Sedra, and Weinberger]{huang2016deep}
Gao Huang, Yu Sun, Zhuang Liu, Daniel Sedra, and Kilian~Q Weinberger.
\newblock Deep networks with stochastic depth.
\newblock In \emph{Computer Vision--ECCV 2016: 14th European Conference, Amsterdam, The Netherlands, October 11--14, 2016, Proceedings, Part IV 14}, pages 646--661. Springer, 2016.

\bibitem[Huang et~al.(2017)Huang, Liu, Van Der~Maaten, and Weinberger]{huang2017densely}
Gao Huang, Zhuang Liu, Laurens Van Der~Maaten, and Kilian~Q Weinberger.
\newblock Densely connected convolutional networks.
\newblock In \emph{Proceedings of the IEEE conference on computer vision and pattern recognition}, pages 4700--4708, 2017.

\bibitem[Huh et~al.(2024)Huh, Cheung, Wang, and Isola]{huh2024platonic}
Minyoung Huh, Brian Cheung, Tongzhou Wang, and Phillip Isola.
\newblock The platonic representation hypothesis.
\newblock \emph{arXiv preprint arXiv:2405.07987}, 2024.

\bibitem[Jiang et~al.(2024)Jiang, Khorram, and Fuxin]{jiang2024comparing}
Mingqi Jiang, Saeed Khorram, and Li Fuxin.
\newblock Comparing the decision-making mechanisms by transformers and cnns via explanation methods.
\newblock In \emph{Proceedings of the IEEE/CVF Conference on Computer Vision and Pattern Recognition}, pages 9546--9555, 2024.

\bibitem[Kaddour et~al.(2024)Kaddour, Key, Nawrot, Minervini, and Kusner]{kaddour2024no}
Jean Kaddour, Oscar Key, Piotr Nawrot, Pasquale Minervini, and Matt~J Kusner.
\newblock No train no gain: Revisiting efficient training algorithms for transformer-based language models.
\newblock \emph{Advances in Neural Information Processing Systems}, 36, 2024.

\bibitem[Khan et~al.(2023)Khan, Nawaz, and Dengel]{khan2023localized}
Abdul~Hannan Khan, Mohammed~Shariq Nawaz, and Andreas Dengel.
\newblock Localized semantic feature mixers for efficient pedestrian detection in autonomous driving.
\newblock In \emph{Proceedings of the IEEE/CVF Conference on Computer Vision and Pattern Recognition}, pages 5476--5485, 2023.

\bibitem[Kim et~al.(2023{\natexlab{a}})Kim, Park, Kim, Lee, Song, Kim, Kim, Kim, Lee, Kim, et~al.]{kim2023solar}
Dahyun Kim, Chanjun Park, Sanghoon Kim, Wonsung Lee, Wonho Song, Yunsu Kim, Hyeonwoo Kim, Yungi Kim, Hyeonju Lee, Jihoo Kim, et~al.
\newblock Solar 10.7 b: Scaling large language models with simple yet effective depth up-scaling.
\newblock \emph{arXiv preprint arXiv:2312.15166}, 2023{\natexlab{a}}.

\bibitem[Kim et~al.(2024{\natexlab{a}})Kim, Heo, and Han]{kim2024densenets}
Donghyun Kim, Byeongho Heo, and Dongyoon Han.
\newblock Densenets reloaded: Paradigm shift beyond resnets and vits.
\newblock \emph{arXiv preprint arXiv:2403.19588}, 2024{\natexlab{a}}.

\bibitem[Kim et~al.(2024{\natexlab{b}})Kim, Valentino, and Freitas]{kim2024mechanistic}
Geonhee Kim, Marco Valentino, and Andr{\'e} Freitas.
\newblock A mechanistic interpretation of syllogistic reasoning in auto-regressive language models.
\newblock \emph{arXiv preprint arXiv:2408.08590}, 2024{\natexlab{b}}.

\bibitem[Kim et~al.(2023{\natexlab{b}})Kim, Cho, Jung, and Yoon]{kim2023feature}
Woo~Jae Kim, Yoonki Cho, Junsik Jung, and Sung-Eui Yoon.
\newblock Feature separation and recalibration for adversarial robustness.
\newblock In \emph{Proceedings of the IEEE/CVF Conference on Computer Vision and Pattern Recognition}, pages 8183--8192, 2023{\natexlab{b}}.

\bibitem[Kingma(2014)]{kingma2014adam}
Diederik~P Kingma.
\newblock Adam: A method for stochastic optimization.
\newblock \emph{arXiv preprint arXiv:1412.6980}, 2014.

\bibitem[Kobayashi(2024)]{kobayashi2024mean}
Takumi Kobayashi.
\newblock Mean-shift feature transformer.
\newblock In \emph{Proceedings of the IEEE/CVF Conference on Computer Vision and Pattern Recognition}, pages 6047--6056, 2024.

\bibitem[Krizhevsky et~al.(2009)Krizhevsky, Hinton, et~al.]{krizhevsky2009learning}
Alex Krizhevsky, Geoffrey Hinton, et~al.
\newblock Learning multiple layers of features from tiny images.
\newblock 2009.

\bibitem[LeCun et~al.(1998)LeCun, Bottou, Bengio, and Haffner]{lecun1998gradient}
Yann LeCun, L{\'e}on Bottou, Yoshua Bengio, and Patrick Haffner.
\newblock Gradient-based learning applied to document recognition.
\newblock \emph{Proceedings of the IEEE}, 86\penalty0 (11):\penalty0 2278--2324, 1998.

\bibitem[Lee et~al.(2023)Lee, Lee, Kim, Choi, Yoo, and Kim]{lee2023fix}
Dongyeun Lee, Jae~Young Lee, Doyeon Kim, Jaehyun Choi, Jaejun Yoo, and Junmo Kim.
\newblock Fix the noise: Disentangling source feature for controllable domain translation.
\newblock In \emph{Proceedings of the IEEE/CVF Conference on Computer Vision and Pattern Recognition}, pages 14224--14234, 2023.

\bibitem[Li et~al.(2022)Li, Zhuang, Wang, Liang, Chang, and Yang]{li2022automated}
Changlin Li, Bohan Zhuang, Guangrun Wang, Xiaodan Liang, Xiaojun Chang, and Yi Yang.
\newblock Automated progressive learning for efficient training of vision transformers.
\newblock In \emph{Proceedings of the IEEE/CVF Conference on Computer Vision and Pattern Recognition}, pages 12486--12496, 2022.

\bibitem[Li et~al.(2024{\natexlab{a}})Li, Duggal, Singh, Kundu, Shuai, and Wu]{li2024robustness}
Guangrui Li, Rahul Duggal, Aaditya Singh, Kaustav Kundu, Bing Shuai, and Jon Wu.
\newblock Robustness preserving fine-tuning using neuron importance.
\newblock 2024{\natexlab{a}}.

\bibitem[Li et~al.(2024{\natexlab{b}})Li, Cao, Huang, and Chen]{li2024formalityfavoredunravelinglearning}
Jiahuan Li, Yiqing Cao, Shujian Huang, and Jiajun Chen.
\newblock Formality is favored: Unraveling the learning preferences of large language models on data with conflicting knowledge, 2024{\natexlab{b}}.

\bibitem[Li et~al.(2020)Li, Zhang, Li, and Fu]{li2020adversarial}
Kai Li, Yulun Zhang, Kunpeng Li, and Yun Fu.
\newblock Adversarial feature hallucination networks for few-shot learning.
\newblock In \emph{Proceedings of the IEEE/CVF conference on computer vision and pattern recognition}, pages 13470--13479, 2020.

\bibitem[Liang et~al.(2024)Liang, Feng, Zhou, Zhang, Zou, and Bai]{liang2024parameterefficientfinetuningspectraldomain}
Dingkang Liang, Tianrui Feng, Xin Zhou, Yumeng Zhang, Zhikang Zou, and Xiang Bai.
\newblock Parameter-efficient fine-tuning in spectral domain for point cloud learning, 2024.

\bibitem[Liang et~al.(2022)Liang, Zhang, Kwon, Yeung, and Zou]{liang2022mind}
Victor~Weixin Liang, Yuhui Zhang, Yongchan Kwon, Serena Yeung, and James~Y Zou.
\newblock Mind the gap: Understanding the modality gap in multi-modal contrastive representation learning.
\newblock \emph{Advances in Neural Information Processing Systems}, 35:\penalty0 17612--17625, 2022.

\bibitem[Lin et~al.(2023)Lin, Shao, Qian, Pan, Guo, and Liu]{lin2023class}
Haowei Lin, Yijia Shao, Weinan Qian, Ningxin Pan, Yiduo Guo, and Bing Liu.
\newblock Class incremental learning via likelihood ratio based task prediction.
\newblock \emph{arXiv preprint arXiv:2309.15048}, 2023.

\bibitem[Liu et~al.(2024{\natexlab{a}})Liu, Chao, Zhang, Wu, Li, Luu, and Bing]{liu2024zero}
Chaoqun Liu, Qin Chao, Wenxuan Zhang, Xiaobao Wu, Boyang Li, Anh~Tuan Luu, and Lidong Bing.
\newblock Zero-to-strong generalization: Eliciting strong capabilities of large language models iteratively without gold labels.
\newblock \emph{arXiv preprint arXiv:2409.12425}, 2024{\natexlab{a}}.

\bibitem[Liu et~al.(2023)Liu, Yan, Leal-Taix{\'e}, Hays, and Ramanan]{liu2023soft}
Yang Liu, Shen Yan, Laura Leal-Taix{\'e}, James Hays, and Deva Ramanan.
\newblock Soft augmentation for image classification.
\newblock In \emph{Proceedings of the IEEE/CVF Conference on Computer Vision and Pattern Recognition}, pages 16241--16250, 2023.

\bibitem[Liu et~al.(2024{\natexlab{b}})Liu, Yang, Chen, Zhang, Wang, Du, and Du]{liu2024pat}
Yijiang Liu, Huanrui Yang, Youxin Chen, Rongyu Zhang, Miao Wang, Yuan Du, and Li Du.
\newblock Pat: Pruning-aware tuning for large language models.
\newblock \emph{arXiv preprint arXiv:2408.14721}, 2024{\natexlab{b}}.

\bibitem[Lizzo and Heck(2024)]{lizzo2024unlearn}
Tyler Lizzo and Larry Heck.
\newblock Unlearn efficient removal of knowledge in large language models.
\newblock \emph{arXiv preprint arXiv:2408.04140}, 2024.

\bibitem[Loshchilov(2017)]{loshchilov2017decoupled}
I Loshchilov.
\newblock Decoupled weight decay regularization.
\newblock \emph{arXiv preprint arXiv:1711.05101}, 2017.

\bibitem[Luo et~al.(2024)Luo, Yang, Dou, Wang, Dai, Qiao, and Zhu]{luo2024monointernvlpushingboundariesmonolithic}
Gen Luo, Xue Yang, Wenhan Dou, Zhaokai Wang, Jifeng Dai, Yu Qiao, and Xizhou Zhu.
\newblock Mono-internvl: Pushing the boundaries of monolithic multimodal large language models with endogenous visual pre-training, 2024.

\bibitem[Lv et~al.(2024)Lv, Xie, Sun, Kang, and Yan]{lv2024language}
Ang Lv, Ruobing Xie, Xingwu Sun, Zhanhui Kang, and Rui Yan.
\newblock Language models" grok" to copy.
\newblock \emph{arXiv preprint arXiv:2409.09281}, 2024.

\bibitem[Ma et~al.(2023)Ma, Jiao, Liu, Yang, Liu, and Li]{ma2023curvature}
Yanbiao Ma, Licheng Jiao, Fang Liu, Shuyuan Yang, Xu Liu, and Lingling Li.
\newblock Curvature-balanced feature manifold learning for long-tailed classification.
\newblock In \emph{Proceedings of the IEEE/CVF conference on computer vision and pattern recognition}, pages 15824--15835, 2023.

\bibitem[Mentzer et~al.(2023)Mentzer, Minnen, Agustsson, and Tschannen]{mentzer2023finite}
Fabian Mentzer, David Minnen, Eirikur Agustsson, and Michael Tschannen.
\newblock Finite scalar quantization: Vq-vae made simple.
\newblock \emph{arXiv preprint arXiv:2309.15505}, 2023.

\bibitem[Ni et~al.(2024)Ni, Zhang, and Koniusz]{ni2024pace}
Yao Ni, Shan Zhang, and Piotr Koniusz.
\newblock Pace: marrying generalization in parameter-efficient fine-tuning with consistency regularization.
\newblock \emph{arXiv preprint arXiv:2409.17137}, 2024.

\bibitem[Oh et~al.(2024)Oh, Shin, and Oh]{oh2024house}
Jaehoon Oh, Seungjun Shin, and Dokwan Oh.
\newblock House of cards: Massive weights in llms.
\newblock \emph{arXiv preprint arXiv:2410.01866}, 2024.

\bibitem[Olah et~al.(2020)Olah, Cammarata, Schubert, Goh, Petrov, and Carter]{olah2020zoom}
Chris Olah, Nick Cammarata, Ludwig Schubert, Gabriel Goh, Michael Petrov, and Shan Carter.
\newblock Zoom in: An introduction to circuits.
\newblock \emph{Distill}, 5\penalty0 (3):\penalty0 e00024--001, 2020.

\bibitem[OpenAI et~al.(2024)OpenAI, Achiam, Adler, Agarwal, Ahmad, Akkaya, Aleman, Almeida, Altenschmidt, Altman, Anadkat, Avila, Babuschkin, Balaji, Balcom, Baltescu, Bao, Bavarian, Belgum, Bello, Berdine, Bernadett-Shapiro, Berner, Bogdonoff, Boiko, Boyd, Brakman, Brockman, Brooks, Brundage, Button, Cai, Campbell, Cann, Carey, Carlson, Carmichael, Chan, Chang, Chantzis, Chen, Chen, Chen, Chen, Chen, Chess, Cho, Chu, Chung, Cummings, Currier, Dai, Decareaux, Degry, Deutsch, Deville, Dhar, Dohan, Dowling, Dunning, Ecoffet, Eleti, Eloundou, Farhi, Fedus, Felix, Fishman, Forte, Fulford, Gao, Georges, Gibson, Goel, Gogineni, Goh, Gontijo-Lopes, Gordon, Grafstein, Gray, Greene, Gross, Gu, Guo, Hallacy, Han, Harris, He, Heaton, Heidecke, Hesse, Hickey, Hickey, Hoeschele, Houghton, Hsu, Hu, Hu, Huizinga, Jain, Jain, Jang, Jiang, Jiang, Jin, Jin, Jomoto, Jonn, Jun, Kaftan, Łukasz Kaiser, Kamali, Kanitscheider, Keskar, Khan, Kilpatrick, Kim, Kim, Kim, Kirchner, Kiros, Knight, Kokotajlo, Łukasz Kondraciuk, Kondrich,
  Konstantinidis, Kosic, Krueger, Kuo, Lampe, Lan, Lee, Leike, Leung, Levy, Li, Lim, Lin, Lin, Litwin, Lopez, Lowe, Lue, Makanju, Malfacini, Manning, Markov, Markovski, Martin, Mayer, Mayne, McGrew, McKinney, McLeavey, McMillan, McNeil, Medina, Mehta, Menick, Metz, Mishchenko, Mishkin, Monaco, Morikawa, Mossing, Mu, Murati, Murk, Mély, Nair, Nakano, Nayak, Neelakantan, Ngo, Noh, Ouyang, O'Keefe, Pachocki, Paino, Palermo, Pantuliano, Parascandolo, Parish, Parparita, Passos, Pavlov, Peng, Perelman, de~Avila Belbute~Peres, Petrov, de~Oliveira~Pinto, Michael, Pokorny, Pokrass, Pong, Powell, Power, Power, Proehl, Puri, Radford, Rae, Ramesh, Raymond, Real, Rimbach, Ross, Rotsted, Roussez, Ryder, Saltarelli, Sanders, Santurkar, Sastry, Schmidt, Schnurr, Schulman, Selsam, Sheppard, Sherbakov, Shieh, Shoker, Shyam, Sidor, Sigler, Simens, Sitkin, Slama, Sohl, Sokolowsky, Song, Staudacher, Such, Summers, Sutskever, Tang, Tezak, Thompson, Tillet, Tootoonchian, Tseng, Tuggle, Turley, Tworek, Uribe, Vallone, Vijayvergiya,
  Voss, Wainwright, Wang, Wang, Wang, Ward, Wei, Weinmann, Welihinda, Welinder, Weng, Weng, Wiethoff, Willner, Winter, Wolrich, Wong, Workman, Wu, Wu, Wu, Xiao, Xu, Yoo, Yu, Yuan, Zaremba, Zellers, Zhang, Zhang, Zhao, Zheng, Zhuang, Zhuk, and Zoph]{openai2024gpt4technicalreport}
OpenAI, Josh Achiam, Steven Adler, Sandhini Agarwal, Lama Ahmad, Ilge Akkaya, Florencia~Leoni Aleman, Diogo Almeida, Janko Altenschmidt, Sam Altman, Shyamal Anadkat, Red Avila, Igor Babuschkin, Suchir Balaji, Valerie Balcom, Paul Baltescu, Haiming Bao, Mohammad Bavarian, Jeff Belgum, Irwan Bello, Jake Berdine, Gabriel Bernadett-Shapiro, Christopher Berner, Lenny Bogdonoff, Oleg Boiko, Madelaine Boyd, Anna-Luisa Brakman, Greg Brockman, Tim Brooks, Miles Brundage, Kevin Button, Trevor Cai, Rosie Campbell, Andrew Cann, Brittany Carey, Chelsea Carlson, Rory Carmichael, Brooke Chan, Che Chang, Fotis Chantzis, Derek Chen, Sully Chen, Ruby Chen, Jason Chen, Mark Chen, Ben Chess, Chester Cho, Casey Chu, Hyung~Won Chung, Dave Cummings, Jeremiah Currier, Yunxing Dai, Cory Decareaux, Thomas Degry, Noah Deutsch, Damien Deville, Arka Dhar, David Dohan, Steve Dowling, Sheila Dunning, Adrien Ecoffet, Atty Eleti, Tyna Eloundou, David Farhi, Liam Fedus, Niko Felix, Simón~Posada Fishman, Juston Forte, Isabella Fulford, Leo
  Gao, Elie Georges, Christian Gibson, Vik Goel, Tarun Gogineni, Gabriel Goh, Rapha Gontijo-Lopes, Jonathan Gordon, Morgan Grafstein, Scott Gray, Ryan Greene, Joshua Gross, Shixiang~Shane Gu, Yufei Guo, Chris Hallacy, Jesse Han, Jeff Harris, Yuchen He, Mike Heaton, Johannes Heidecke, Chris Hesse, Alan Hickey, Wade Hickey, Peter Hoeschele, Brandon Houghton, Kenny Hsu, Shengli Hu, Xin Hu, Joost Huizinga, Shantanu Jain, Shawn Jain, Joanne Jang, Angela Jiang, Roger Jiang, Haozhun Jin, Denny Jin, Shino Jomoto, Billie Jonn, Heewoo Jun, Tomer Kaftan, Łukasz Kaiser, Ali Kamali, Ingmar Kanitscheider, Nitish~Shirish Keskar, Tabarak Khan, Logan Kilpatrick, Jong~Wook Kim, Christina Kim, Yongjik Kim, Jan~Hendrik Kirchner, Jamie Kiros, Matt Knight, Daniel Kokotajlo, Łukasz Kondraciuk, Andrew Kondrich, Aris Konstantinidis, Kyle Kosic, Gretchen Krueger, Vishal Kuo, Michael Lampe, Ikai Lan, Teddy Lee, Jan Leike, Jade Leung, Daniel Levy, Chak~Ming Li, Rachel Lim, Molly Lin, Stephanie Lin, Mateusz Litwin, Theresa Lopez, Ryan
  Lowe, Patricia Lue, Anna Makanju, Kim Malfacini, Sam Manning, Todor Markov, Yaniv Markovski, Bianca Martin, Katie Mayer, Andrew Mayne, Bob McGrew, Scott~Mayer McKinney, Christine McLeavey, Paul McMillan, Jake McNeil, David Medina, Aalok Mehta, Jacob Menick, Luke Metz, Andrey Mishchenko, Pamela Mishkin, Vinnie Monaco, Evan Morikawa, Daniel Mossing, Tong Mu, Mira Murati, Oleg Murk, David Mély, Ashvin Nair, Reiichiro Nakano, Rajeev Nayak, Arvind Neelakantan, Richard Ngo, Hyeonwoo Noh, Long Ouyang, Cullen O'Keefe, Jakub Pachocki, Alex Paino, Joe Palermo, Ashley Pantuliano, Giambattista Parascandolo, Joel Parish, Emy Parparita, Alex Passos, Mikhail Pavlov, Andrew Peng, Adam Perelman, Filipe de Avila Belbute~Peres, Michael Petrov, Henrique~Ponde de Oliveira~Pinto, Michael, Pokorny, Michelle Pokrass, Vitchyr~H. Pong, Tolly Powell, Alethea Power, Boris Power, Elizabeth Proehl, Raul Puri, Alec Radford, Jack Rae, Aditya Ramesh, Cameron Raymond, Francis Real, Kendra Rimbach, Carl Ross, Bob Rotsted, Henri Roussez,
  Nick Ryder, Mario Saltarelli, Ted Sanders, Shibani Santurkar, Girish Sastry, Heather Schmidt, David Schnurr, John Schulman, Daniel Selsam, Kyla Sheppard, Toki Sherbakov, Jessica Shieh, Sarah Shoker, Pranav Shyam, Szymon Sidor, Eric Sigler, Maddie Simens, Jordan Sitkin, Katarina Slama, Ian Sohl, Benjamin Sokolowsky, Yang Song, Natalie Staudacher, Felipe~Petroski Such, Natalie Summers, Ilya Sutskever, Jie Tang, Nikolas Tezak, Madeleine~B. Thompson, Phil Tillet, Amin Tootoonchian, Elizabeth Tseng, Preston Tuggle, Nick Turley, Jerry Tworek, Juan Felipe~Cerón Uribe, Andrea Vallone, Arun Vijayvergiya, Chelsea Voss, Carroll Wainwright, Justin~Jay Wang, Alvin Wang, Ben Wang, Jonathan Ward, Jason Wei, CJ Weinmann, Akila Welihinda, Peter Welinder, Jiayi Weng, Lilian Weng, Matt Wiethoff, Dave Willner, Clemens Winter, Samuel Wolrich, Hannah Wong, Lauren Workman, Sherwin Wu, Jeff Wu, Michael Wu, Kai Xiao, Tao Xu, Sarah Yoo, Kevin Yu, Qiming Yuan, Wojciech Zaremba, Rowan Zellers, Chong Zhang, Marvin Zhang, Shengjia
  Zhao, Tianhao Zheng, Juntang Zhuang, William Zhuk, and Barret Zoph.
\newblock Gpt-4 technical report, 2024.

\bibitem[Oquab et~al.(2023)Oquab, Darcet, Moutakanni, Vo, Szafraniec, Khalidov, Fernandez, Haziza, Massa, El-Nouby, et~al.]{oquab2023dinov2}
Maxime Oquab, Timoth{\'e}e Darcet, Th{\'e}o Moutakanni, Huy Vo, Marc Szafraniec, Vasil Khalidov, Pierre Fernandez, Daniel Haziza, Francisco Massa, Alaaeldin El-Nouby, et~al.
\newblock Dinov2: Learning robust visual features without supervision.
\newblock \emph{arXiv preprint arXiv:2304.07193}, 2023.

\bibitem[Park(2024{\natexlab{a}})]{park2024diverse}
Sejik Park.
\newblock Diverse feature learning by self-distillation and reset.
\newblock \emph{arXiv preprint arXiv:2403.19941}, 2024{\natexlab{a}}.

\bibitem[Park(2024{\natexlab{b}})]{park2024learning}
Sejik Park.
\newblock Learning more generalized experts by merging experts in mixture-of-experts.
\newblock \emph{arXiv preprint arXiv:2405.11530}, 2024{\natexlab{b}}.

\bibitem[Podell et~al.(2023)Podell, English, Lacey, Blattmann, Dockhorn, Müller, Penna, and Rombach]{podell2023sdxlimprovinglatentdiffusion}
Dustin Podell, Zion English, Kyle Lacey, Andreas Blattmann, Tim Dockhorn, Jonas Müller, Joe Penna, and Robin Rombach.
\newblock Sdxl: Improving latent diffusion models for high-resolution image synthesis, 2023.

\bibitem[Potje et~al.(2023)Potje, Cadar, Araujo, Martins, and Nascimento]{potje2023enhancing}
Guilherme Potje, Felipe Cadar, Andr{\'e} Araujo, Renato Martins, and Erickson~R Nascimento.
\newblock Enhancing deformable local features by jointly learning to detect and describe keypoints.
\newblock In \emph{Proceedings of the IEEE/CVF Conference on Computer Vision and Pattern Recognition}, pages 1306--1315, 2023.

\bibitem[Qiu et~al.(2023)Qiu, Li, Wen, Qiu, Wang, Meng, Wu, and Pan]{qiu2023cafeboost}
Benliu Qiu, Hongliang Li, Haitao Wen, Heqian Qiu, Lanxiao Wang, Fanman Meng, Qingbo Wu, and Lili Pan.
\newblock Cafeboost: Causal feature boost to eliminate task-induced bias for class incremental learning.
\newblock In \emph{Proceedings of the IEEE/CVF Conference on Computer Vision and Pattern Recognition}, pages 16016--16025, 2023.

\bibitem[Qiu et~al.(2024)Qiu, Huang, and Fu]{qiu2024unlocking}
Zihan Qiu, Zeyu Huang, and Jie Fu.
\newblock Unlocking emergent modularity in large language models.
\newblock In \emph{Proceedings of the 2024 Conference of the North American Chapter of the Association for Computational Linguistics: Human Language Technologies (Volume 1: Long Papers)}, pages 2638--2660, 2024.

\bibitem[Russakovsky et~al.(2015)Russakovsky, Deng, Su, Krause, Satheesh, Ma, Huang, Karpathy, Khosla, Bernstein, Berg, and Fei-Fei]{imagenet15russakovsky}
Olga Russakovsky, Jia Deng, Hao Su, Jonathan Krause, Sanjeev Satheesh, Sean Ma, Zhiheng Huang, Andrej Karpathy, Aditya Khosla, Michael Bernstein, Alexander~C. Berg, and Li Fei-Fei.
\newblock {ImageNet Large Scale Visual Recognition Challenge}.
\newblock \emph{International Journal of Computer Vision (IJCV)}, 115\penalty0 (3):\penalty0 211--252, 2015.

\bibitem[Sakurai et~al.(2024)Sakurai, Ishii, Shimizu, Song, and Goto]{sakurai2024lare}
Kosuke Sakurai, Tatsuya Ishii, Ryotaro Shimizu, Linxin Song, and Masayuki Goto.
\newblock Lare: Latent augmentation using regional embedding with vision-language model.
\newblock \emph{arXiv preprint arXiv:2409.12597}, 2024.

\bibitem[Samragh et~al.(2024)Samragh, Mirzadeh, Vahid, Faghri, Cho, Nabi, Naik, and Farajtabar]{samragh2024scaling}
Mohammad Samragh, Iman Mirzadeh, Keivan~Alizadeh Vahid, Fartash Faghri, Minsik Cho, Moin Nabi, Devang Naik, and Mehrdad Farajtabar.
\newblock Scaling smart: Accelerating large language model pre-training with small model initialization.
\newblock \emph{arXiv preprint arXiv:2409.12903}, 2024.

\bibitem[Serra et~al.(2018)Serra, Suris, Miron, and Karatzoglou]{serra2018overcoming}
Joan Serra, Didac Suris, Marius Miron, and Alexandros Karatzoglou.
\newblock Overcoming catastrophic forgetting with hard attention to the task.
\newblock In \emph{International conference on machine learning}, pages 4548--4557. PMLR, 2018.

\bibitem[Simon et~al.(2020)Simon, Koniusz, Nock, and Harandi]{simon2020adaptive}
Christian Simon, Piotr Koniusz, Richard Nock, and Mehrtash Harandi.
\newblock Adaptive subspaces for few-shot learning.
\newblock In \emph{Proceedings of the IEEE/CVF conference on computer vision and pattern recognition}, pages 4136--4145, 2020.

\bibitem[Son et~al.(2024)Son, Choi, and Min]{son2024sg}
Sumin Son, Hyesong Choi, and Dongbo Min.
\newblock Sg-mim: Structured knowledge guided efficient pre-training for dense prediction.
\newblock \emph{arXiv preprint arXiv:2409.02513}, 2024.

\bibitem[Song et~al.(2024)Song, Xu, and Zhong]{song2024out}
Jiajun Song, Zhuoyan Xu, and Yiqiao Zhong.
\newblock Out-of-distribution generalization via composition: a lens through induction heads in transformers.
\newblock \emph{arXiv preprint arXiv:2408.09503}, 2024.

\bibitem[Springenberg et~al.(2014)Springenberg, Dosovitskiy, Brox, and Riedmiller]{springenberg2014striving}
Jost~Tobias Springenberg, Alexey Dosovitskiy, Thomas Brox, and Martin Riedmiller.
\newblock Striving for simplicity: The all convolutional net.
\newblock \emph{arXiv preprint arXiv:1412.6806}, 2014.

\bibitem[Srivastava et~al.(2014)Srivastava, Hinton, Krizhevsky, Sutskever, and Salakhutdinov]{srivastava2014dropout}
Nitish Srivastava, Geoffrey Hinton, Alex Krizhevsky, Ilya Sutskever, and Ruslan Salakhutdinov.
\newblock Dropout: a simple way to prevent neural networks from overfitting.
\newblock \emph{The journal of machine learning research}, 15\penalty0 (1):\penalty0 1929--1958, 2014.

\bibitem[Staats et~al.(2024)Staats, Thamm, and Rosenow]{staats2024locating}
Max Staats, Matthias Thamm, and Bernd Rosenow.
\newblock Locating information in large language models via random matrix theory.
\newblock \emph{arXiv preprint arXiv:2410.17770}, 2024.

\bibitem[Sun et~al.(2024)Sun, Wei, Wu, Shi, He, Ma, Xie, and Yang]{sun2024svfit}
Chengwei Sun, Jiwei Wei, Yujia Wu, Yiming Shi, Shiyuan He, Zeyu Ma, Ning Xie, and Yang Yang.
\newblock Svfit: Parameter-efficient fine-tuning of large pre-trained models using singular values.
\newblock \emph{arXiv preprint arXiv:2409.05926}, 2024.

\bibitem[Tang et~al.(2024)Tang, Shen, Luo, Xie, Hu, Zhang, Du, and Tao]{tang2024smile}
Anke Tang, Li Shen, Yong Luo, Shuai Xie, Han Hu, Lefei Zhang, Bo Du, and Dacheng Tao.
\newblock Smile: Zero-shot sparse mixture of low-rank experts construction from pre-trained foundation models.
\newblock \emph{arXiv preprint arXiv:2408.10174}, 2024.

\bibitem[Templeton(2024)]{templeton2024scaling}
Adly Templeton.
\newblock \emph{Scaling monosemanticity: Extracting interpretable features from claude 3 sonnet}.
\newblock Anthropic, 2024.

\bibitem[Touvron et~al.(2022)Touvron, Cord, and J{\'e}gou]{touvron2022deit}
Hugo Touvron, Matthieu Cord, and Herv{\'e} J{\'e}gou.
\newblock Deit iii: Revenge of the vit.
\newblock In \emph{European conference on computer vision}, pages 516--533. Springer, 2022.

\bibitem[Touvron et~al.(2023)Touvron, Martin, Stone, Albert, Almahairi, Babaei, Bashlykov, Batra, Bhargava, Bhosale, et~al.]{touvron2023llama}
Hugo Touvron, Louis Martin, Kevin Stone, Peter Albert, Amjad Almahairi, Yasmine Babaei, Nikolay Bashlykov, Soumya Batra, Prajjwal Bhargava, Shruti Bhosale, et~al.
\newblock Llama 2: Open foundation and fine-tuned chat models.
\newblock \emph{arXiv preprint arXiv:2307.09288}, 2023.

\bibitem[Tran et~al.(2024)Tran, Thanh, Anh, Hai, Le, Van~Ngo, and Nguyen]{tran2024preserving}
Quyen Tran, Nguyen~Xuan Thanh, Nguyen~Hoang Anh, Nam~Le Hai, Trung Le, Linh Van~Ngo, and Thien~Huu Nguyen.
\newblock Preserving generalization of language models in few-shot continual relation extraction.
\newblock \emph{arXiv preprint arXiv:2410.00334}, 2024.

\bibitem[Tseng et~al.(2020)Tseng, Chen, Tsai, Liu, Lin, and Yang]{tseng2020regularizing}
Hung-Yu Tseng, Yi-Wen Chen, Yi-Hsuan Tsai, Sifei Liu, Yen-Yu Lin, and Ming-Hsuan Yang.
\newblock Regularizing meta-learning via gradient dropout.
\newblock In \emph{Proceedings of the Asian Conference on Computer Vision}, 2020.

\bibitem[Uchiyama et~al.(2024)Uchiyama, Kojima, Gambardella, Cao, Iwasawa, and Matsuo]{uchiyama2024programminglanguagefeaturespretraining}
Fumiya Uchiyama, Takeshi Kojima, Andrew Gambardella, Qi Cao, Yusuke Iwasawa, and Yutaka Matsuo.
\newblock Which programming language and what features at pre-training stage affect downstream logical inference performance?, 2024.

\bibitem[Vesdapunt et~al.(2024)Vesdapunt, Fu, Wu, Zhang, and Natarajan]{vesdapunt2024hvclip}
Sol Vesdapunt, Kah~Kuen Fu, Yue~Rex Wu, Xu Zhang, and Pradeep Natarajan.
\newblock Hvclip: High-dimensional vector in clip for unsupervised domain adaptation.
\newblock 2024.

\bibitem[Wang et~al.(2023{\natexlab{a}})Wang, Li, Nakashima, and Nagahara]{wang2023learning}
Bowen Wang, Liangzhi Li, Yuta Nakashima, and Hajime Nagahara.
\newblock Learning bottleneck concepts in image classification.
\newblock In \emph{Proceedings of the ieee/cvf conference on computer vision and pattern recognition}, pages 10962--10971, 2023{\natexlab{a}}.

\bibitem[Wang et~al.(2023{\natexlab{b}})Wang, Niu, Zhang, and Zhang]{wang2023image}
Chao Wang, Li Niu, Bo Zhang, and Liqing Zhang.
\newblock Image cropping with spatial-aware feature and rank consistency.
\newblock In \emph{Proceedings of the IEEE/CVF Conference on Computer Vision and Pattern Recognition}, pages 10052--10061, 2023{\natexlab{b}}.

\bibitem[Wang et~al.(2024)Wang, Yeh, and Liao]{wang2024yolov9}
Chien-Yao Wang, I-Hau Yeh, and Hong-Yuan~Mark Liao.
\newblock Yolov9: Learning what you want to learn using programmable gradient information.
\newblock \emph{arXiv preprint arXiv:2402.13616}, 2024.

\bibitem[Wang et~al.(2023{\natexlab{c}})Wang, Yue, Lu, Liu, Zhong, Song, and Huang]{wang2023efficienttrain}
Yulin Wang, Yang Yue, Rui Lu, Tianjiao Liu, Zhao Zhong, Shiji Song, and Gao Huang.
\newblock Efficienttrain: Exploring generalized curriculum learning for training visual backbones.
\newblock In \emph{Proceedings of the IEEE/CVF International Conference on Computer Vision}, pages 5852--5864, 2023{\natexlab{c}}.

\bibitem[Warmuth et~al.(2020)Warmuth, Kot{\l}owski, and Amid]{warmuth2020case}
Manfred~K Warmuth, Wojciech Kot{\l}owski, and Ehsan Amid.
\newblock A case where a spindly two-layer linear network whips any neural network with a fully connected input layer.
\newblock \emph{arXiv preprint arXiv:2010.08625}, 2020.

\bibitem[Wightman(2019)]{rw2019timm}
Ross Wightman.
\newblock Pytorch image models.
\newblock \url{https://github.com/rwightman/pytorch-image-models}, 2019.

\bibitem[Wu et~al.(2024)Wu, Zhang, Deng, Duan, and Deng]{wu2024panadapter}
RuoCheng Wu, ZiEn Zhang, ShangQi Deng, YuLe Duan, and LiangJian Deng.
\newblock Panadapter: Two-stage fine-tuning with spatial-spectral priors injecting for pansharpening.
\newblock \emph{arXiv preprint arXiv:2409.06980}, 2024.

\bibitem[Xia et~al.()Xia, Wang, Lv, Hao, and Shi]{xia2403vit}
C Xia, X Wang, F Lv, X Hao, and Y Shi.
\newblock Vit-comer: Vision transformer with convolutional multi-scale feature interaction for dense predictions. arxiv 2024.
\newblock \emph{arXiv preprint arXiv:2403.07392}.

\bibitem[Xiao et~al.(2024)Xiao, Meng, Li, and Yuan]{xiao2024improving}
Da Xiao, Qingye Meng, Shengping Li, and Xingyuan Yuan.
\newblock Improving transformers with dynamically composable multi-head attention.
\newblock \emph{arXiv preprint arXiv:2405.08553}, 2024.

\bibitem[Xie et~al.(2024)Xie, Wang, Yan, Zhou, Deng, and Huang]{xie2024making}
Yukang Xie, Chengyu Wang, Junbing Yan, Jiyong Zhou, Feiqi Deng, and Jun Huang.
\newblock Making small language models better multi-task learners with mixture-of-task-adapters.
\newblock In \emph{Proceedings of the 17th ACM International Conference on Web Search and Data Mining}, pages 1094--1097, 2024.

\bibitem[Xu et~al.(2024)Xu, Chen, Li, Zhao, and Wei]{xu2024collapsedlanguagemodelspromote}
Jingxuan Xu, Wuyang Chen, Linyi Li, Yao Zhao, and Yunchao Wei.
\newblock Collapsed language models promote fairness, 2024.

\bibitem[Yang et~al.(2023)Yang, Yang, Butt, van~de Weijer, et~al.]{yang2023dynamic}
Fei Yang, Shiqi Yang, Muhammad~Atif Butt, Joost van~de Weijer, et~al.
\newblock Dynamic prompt learning: Addressing cross-attention leakage for text-based image editing.
\newblock \emph{Advances in Neural Information Processing Systems}, 36:\penalty0 26291--26303, 2023.

\bibitem[Yang et~al.(2024)Yang, Han, Wang, and Liu]{yang2024graphlora}
Zhe-Rui Yang, Jindong Han, Chang-Dong Wang, and Hao Liu.
\newblock Graphlora: Structure-aware contrastive low-rank adaptation for cross-graph transfer learning.
\newblock \emph{arXiv preprint arXiv:2409.16670}, 2024.

\bibitem[Yu et~al.(2023{\natexlab{a}})Yu, Chang, He, Zhang, Yu, and Wu]{yu2023adaptive}
Jiahuan Yu, Jiahao Chang, Jianfeng He, Tianzhu Zhang, Jiyang Yu, and Feng Wu.
\newblock Adaptive spot-guided transformer for consistent local feature matching.
\newblock In \emph{Proceedings of the IEEE/CVF Conference on Computer Vision and Pattern Recognition}, pages 21898--21908, 2023{\natexlab{a}}.

\bibitem[Yu et~al.(2023{\natexlab{b}})Yu, Shin, Lee, Jun, and Lee]{yu2023block}
Yeonguk Yu, Sungho Shin, Seongju Lee, Changhyun Jun, and Kyoobin Lee.
\newblock Block selection method for using feature norm in out-of-distribution detection.
\newblock In \emph{Proceedings of the IEEE/CVF Conference on Computer Vision and Pattern Recognition}, pages 15701--15711, 2023{\natexlab{b}}.

\bibitem[Yun et~al.(2019)Yun, Han, Oh, Chun, Choe, and Yoo]{yun2019cutmix}
Sangdoo Yun, Dongyoon Han, Seong~Joon Oh, Sanghyuk Chun, Junsuk Choe, and Youngjoon Yoo.
\newblock Cutmix: Regularization strategy to train strong classifiers with localizable features.
\newblock In \emph{Proceedings of the IEEE/CVF international conference on computer vision}, pages 6023--6032, 2019.

\bibitem[Zeiler(2014)]{zeiler2014visualizing}
MD Zeiler.
\newblock Visualizing and understanding convolutional networks.
\newblock In \emph{European conference on computer vision/arXiv}, 2014.

\bibitem[Zhai et~al.(2023)Zhai, Zhao, Long, Xu, He, and Zhao]{zhai2023feature}
Zhijun Zhai, Jianhui Zhao, Chengjiang Long, Wenju Xu, Shuangjiang He, and Huijuan Zhao.
\newblock Feature representation learning with adaptive displacement generation and transformer fusion for micro-expression recognition.
\newblock In \emph{Proceedings of the IEEE/CVF Conference on Computer Vision and Pattern Recognition}, pages 22086--22095, 2023.

\bibitem[Zhang et~al.(2020)Zhang, Cai, Lin, and Shen]{zhang2020deepemd}
Chi Zhang, Yujun Cai, Guosheng Lin, and Chunhua Shen.
\newblock Deepemd: Few-shot image classification with differentiable earth mover's distance and structured classifiers.
\newblock In \emph{Proceedings of the IEEE/CVF conference on computer vision and pattern recognition}, pages 12203--12213, 2020.

\bibitem[Zhang(2017)]{zhang2017mixup}
Hongyi Zhang.
\newblock mixup: Beyond empirical risk minimization.
\newblock \emph{arXiv preprint arXiv:1710.09412}, 2017.

\bibitem[Zhang et~al.(2024{\natexlab{a}})Zhang, Li, and Wu]{zhang2024co}
Xiao Zhang, Miao Li, and Ji Wu.
\newblock Co-occurrence is not factual association in language models.
\newblock \emph{arXiv preprint arXiv:2409.14057}, 2024{\natexlab{a}}.

\bibitem[Zhang et~al.(2024{\natexlab{b}})Zhang, Sui, and Yeung-Levy]{zhang2024connect}
Yuhui Zhang, Elaine Sui, and Serena Yeung-Levy.
\newblock Connect, collapse, corrupt: Learning cross-modal tasks with uni-modal data.
\newblock \emph{arXiv preprint arXiv:2401.08567}, 2024{\natexlab{b}}.

\bibitem[Zhao et~al.(2024)Zhao, Xu, Xu, Song, Wang, Zhou, and Bian]{zhao2024enhancing}
Wenhao Zhao, Qiushui Xu, Linjie Xu, Lei Song, Jinyu Wang, Chunlai Zhou, and Jiang Bian.
\newblock Enhancing cross-domain pre-trained decision transformers with adaptive attention.
\newblock \emph{arXiv preprint arXiv:2409.06985}, 2024.

\bibitem[Zheng et~al.(2024)Zheng, Yang, and Yeh]{zheng2024learning}
Amber~Yijia Zheng, Chiao-An Yang, and Raymond~A Yeh.
\newblock Learning to obstruct few-shot image classification over restricted classes.
\newblock \emph{arXiv preprint arXiv:2409.19210}, 2024.

\bibitem[Zoph et~al.(2020)Zoph, Cubuk, Ghiasi, Lin, Shlens, and Le]{zoph2020learning}
Barret Zoph, Ekin~D Cubuk, Golnaz Ghiasi, Tsung-Yi Lin, Jonathon Shlens, and Quoc~V Le.
\newblock Learning data augmentation strategies for object detection.
\newblock In \emph{Computer Vision--ECCV 2020: 16th European Conference, Glasgow, UK, August 23--28, 2020, Proceedings, Part XXVII 16}, pages 566--583. Springer, 2020.

\end{thebibliography}
